\documentclass[sigconf]{acmart}

\AtBeginDocument{%
  \providecommand\BibTeX{{%
    \normalfont B\kern-0.5em{\scshape i\kern-0.25em b}\kern-0.8em\TeX}}}

\setcopyright{acmcopyright}
\copyrightyear{2018}
\acmYear{2018}
\acmDOI{XXXXXXX.XXXXXXX}

\acmConference[ICAIF]{ACM International Conference
on AI in Finance}{November, 2023}{NY, NY}
\acmPrice{15.00}
\acmISBN{978-1-4503-XXXX-X/18/06}

\usepackage{algpseudocode}
\usepackage{algorithm}
\usepackage{multirow}
\usepackage{caption}
\usepackage{subcaption}
\usepackage{float}
\usepackage{subcaption}

\copyrightyear{2023}
\acmYear{2023}
\setcopyright{acmlicensed}\acmConference[ICAIF '23]{4th ACM International Conference on AI in Finance}{November 27--29, 2023}{Brooklyn, NY, USA}
\acmBooktitle{4th ACM International Conference on AI in Finance (ICAIF '23), November 27--29, 2023, Brooklyn, NY, USA}
\acmPrice{15.00}
\acmDOI{10.1145/3604237.3626859}
\acmISBN{979-8-4007-0240-2/23/11}
\begin{document}

\title{Conservative Predictions on Noisy Financial Data}

\author{Omkar Nabar}
\affiliation{
\institution{BITS Pilani, Goa}
\country{India}
}

\author{Gautam Shroff}
\affiliation{%
  \institution{TCS Research, Delhi}
  \country{India}
}


\begin{abstract}
Price movements in financial markets are well known to
be very noisy. As a result, even if there are, on occasion,
exploitable patterns that could be picked up by machine-learning
algorithms, these are obscured by feature and label noise 
rendering the predictions less useful, and risky in practice. 
Traditional rule-learning techniques developed for noisy data, 
such as CN2, would seek only high precision rules and refrain from making predictions where their antecedents did not apply. We apply a similar 
approach, where a model abstains from making a prediction on data points
that it is uncertain on. During training, a cascade of such 
models are learned in sequence, similar to rule lists, with
each model being trained only on data on which the previous 
model(s) were uncertain. Similar pruning of data takes place
at test-time, with (higher accuracy) predictions being made 
albeit only on a fraction (support) of test-time data. In
a financial prediction setting, such an approach allows decisions
to be taken only when the ensemble model is confident, thereby
reducing risk. We present results using traditional MLPs
as well as differentiable decision trees, on synthetic data
as well as real financial market data, to predict fixed-term
returns using commonly used features. We submit that
our approach is likely to result in better overall returns at
a lower level of risk. In this context we introduce an \textit{utility}
metric to measure the average gain per trade, as well as the return
adjusted for downside-risk, both of which are improved
significantly by our approach.
\end{abstract}

\maketitle

\section{Introduction}
Machine-learning (especially deep learning) techniques have achieved close to human-level performance
in many domains such as image processing and natural language understanding. However, the same cannot
be said in the case of trading in financial markets: It has been observed that expert human traders consistently 
outperform others, thus in some sense negating the efficient market hypothesis, and also indicating that there 
is expertise involved, presumably at recognizing and acting on recurring patterns, even if ephemeral. 

There have been many reports of attempts to apply machine-learning and deep-learning techniques for prediction
and trading in financial markets, as we shall review later in Section \ref{related_work}. Nevertheless,
in practice these are all plagued by the fact that financial markets are \textit{noisy}. Thus, even if
there are ephemeral patterns in the data that are discovered by machine-learning techniques in spite of
this noise, when applied the models still often result in erroneous predictions, again due to noise. If
one were to believe the efficient market hypothesis, there would be no signal and only noise, and no model 
could succeed consistently.

Some techniques to deal with noisy data in deep-learning have relied on custom loss functions \cite{qin2019making},
\cite{patrini2017making}, as well as training procedures such as bootstrapping' \cite{reed2014training}, 
where data points that train slowly are presumed to be noisy
and their labels replaced with predictions of other similar but likely less noisy points. 

The problem of noise in the data has been recognized historically in machine-learning: The CN2 rule-learning 
algorithm \cite{clark1989cn2} was developed to combat noise by learning a sequence of high accuracy
rules that would apply only on a fraction of data, i.e., some data points would have no applicable rule.
Such cases could be handled in a domain-specific manner, e.g., by predicting the majority class, making
a random choice, or, more appropriate to risky situations such as trading in financial markets, by refraining
from making a prediction. In the context of a trading system relying on such a model, no trade would be taken
if the model refused to make a prediction. 

The relative benefits of relying on simple rules when dealing with financial market data has been highlighted
in \cite{dhar2011prediction}, where it is argued that more complex models only end up fitting the noise in such data.

We are motivated by the following ideas from the above prior works: (i) using simpler rule-based models as advised in \cite{dhar2011prediction}, (ii) learning a sequence of models trained on recursively smaller subsets of data as in CN2
and (iii) refraining from making a prediction any of the sequence of models are inapplicable. 

Our approach
involves learning a \textit{cascade} of models in sequence, with each successive model being trained on data
on which the previous models are under-confident, as measured by the Gini index of the predicted class distribution
for each data point. Cascaded training continues as long as the accuracy obtained on data where the model \textit{is}
confident are above a minimum desired value. We use a three level cascade in our experiments that appears to work
well. Finally, we report accuracy on the remaining `un-pruned' data points as well as the support, i.e., fraction of data
that the model makes confident predictions on. 

Since we found that hard-decision rule-based models as learned by RIPPER performed poorly on our datasets, we employ
differentiable decision trees (DDT) as introduced in \cite{suarez1999fuzzy} and used in a deep-learning context by \cite{silva2020optimization}. We also used traditional MLPs in our cascaded learning approach, and compare these 
with DDTs.

We report experimental results on real market data as well as synthetic data created to mimic a simplistic
mean-reverting market behavior using sine waves. Varying levels of noise are added to this synthetic data to
study whether using a cascade of models improves accuracy with acceptable support.

Note that in a financial market scenario, it is preferable to have a model with 70\% or even 60\% accuracy on
say 20\% or even 10\% of the data, on which it makes confident predictions, and abstains on the balance, as this
serves to minimise risk in any decisions taken based on the model's predictions. Further, it is more useful
to have confident predictions at the extremes of the target attribute's distribution: Such predictions are 
actionable (as opposed to confidently predicting placid market behaviors), and we define \textit{utility} of 
predictions as a metric to measure actionability in the above sense.

Utility measures the average gain per trade; thus, higher utility model recommends fewer, albeit successful,
as well as less risky trades.  We also compute a measure of the return adjusted for downside risk.
Our experimental results show that using a cascade of models indeed achieves this effect especially on synthetic
data, with DDTs resulting in higher support at comparable accuracy to MLPs, and higher utility as well
as risk-adjusted return.
On real-data, while both models perform relatively poorly with respect to the final support, we observe
that their predictions, however rare, nevertheless exhibit higher utility and risk-adjusted return, and are 
therefore actionable with lower level of risk. Finally we discuss how this behaviour can be exploited 
in the context of financial trading strategies.

\section{Methods: Algorithm \& Models}
\subsection{Differentiable Decision Trees (DDT)}
Decision Trees have long been used to generate interpretable models for tabular datasets and perform well with classification problems. However, despite the numerous algorithms developed to produce near optimal decision trees, they still are highly sensitive to their training set. Deep Neural Networks (DNN) or \textbf{Multi-Layer Perceptrons (MLP)}, on the other hand are analytic in nature and thus can be optimised to produce good results on both regression and classification problems. However, this comes at a cost of interpretability. 

Suarez and Lutsko \cite{suarez1999fuzzy} proposed a modification to the `crisp' decision trees formed by the \textbf{CART algorithm}. They replaced the hard decisions taken at each node with `fuzzy' decisions, determined by the sigmoid function. The leaf nodes are modified to represent a probability distribution over the classes. This allows the fuzzy decision tree or `differentiable decision tree' to be trained with gradient descent like a multi-layer perceptron. 

Further work by Frosst and Hinton \cite{frosst2017sdt} introduced a regularization term to encourage a balanced split at each internal nodes. The differentiable decision tree model used in the experiments is a combination of these ideas from previous works.
\subsubsection{Forward pass and optimization}
The decision tree is initialized as a balanced binary tree, with the depth fixed before-hand. The model works in a hierarchical manner,  with each inner node \(i\) having its own weight \(w_{i}\) and bias \(b_{i}\) while the leaf nodes have a learned distribution \(Q_{l}\). At each inner node \(i\), the probability of taking the right branch is given by the sigmoid function:
\[p_{i} = \sigma(xw_{i} + b_{i})\]
To prevent the decisions from being too soft, temperature scaling ($T$) is added to the sigmoid function as follows:
\[p_{i} = \sigma(T(xw_{i} + b_{i}))\]
At each inner node \(i\) the path probability up until that node is given by \(P_{i}\). This value is then multiplied by \(p_{i}\) for the right branch and \(1-p_{i}\) for the left branch which get passed down as the path probability for that node. The final output is calculated as a \textbf{weighted sum} of the probability distributions of all the leaves:
\[p = \Sigma_{l \in leaves} P_{l}Q_{l}\]
where \(P_{l}\) is the path probability at leaf node \(l\). The \textbf{cross-entropy} loss on this output \(p\) is used for the optimization of the model. Applying the softmax function on \(p\) gives us the probability for each class.
\[P_{c} = \frac{e^{-p_{c}}}{\Sigma_{c' \in classes}e^{-p_{c'}}}\]
\(P_{c}\) denotes the probability of class \(c\) while \(p_{c}\)  is the \(c^{th}\) element in the output p. The cross-entropy is minimized by
gradient descent to train the parameters $w_i,b_i$ as well as the scaling parameter $T$.
\subsubsection{Regularization:} 
The DDT tends to get stuck in a local minima quite often, with one path getting much higher probabilities than the others, which results in the model favouring one leaf over the others for most of the predictions. To solve this issue, Frrost and Hinton \cite{frosst2017sdt}
introduced a regularization term which makes the inner nodes make a more balanced split. At each inner node \(i\) an \(\alpha\) term is calculated:
\[\alpha_{i} = \frac{\Sigma_{x}P_{i}(x)p_{i}(x)}{\Sigma_{x}P_{i}(x)}\]
The penalty summed over all the terms is:
\[P =-\lambda \Sigma_{i \in Inner Nodes}0.5 * log(\alpha_{i}) + 0.5 * log(1 - \alpha_{i})\]
where \(\lambda\) varies as \(2^{-d}\) where \(d\) is the depth of the node.

\subsection{Cascading Models}
We introduce a method in which the model makes selective predictions, and chooses to not predict on some part of the data when it's not confident about its prediction. We start with a classification model, which outputs probabilities of the classes, such as an MLP or DDT. The more imbalanced the probabilities, more confident the model is on it's predictions. We use \textbf{Gini Impurity} to calculate the imbalance in the probabilities:
\[Gini  Impurity = 1 - \Sigma_{c \in classes}p^2_{c}\]

The lower the gini impurity, greater the imbalance in the probability distribution of the classes. We set a value to be the maximum admissible impurity in the predictions, and the model is confident on the ones with lower impurity. 

The expectation is that the model would ideally give a much higher accuracy on the subset of the predictions (\textbf{support}) on which it is confident. However, in cases of very noisy data, this support can be very low. Therefore we use \textbf{Cascading Models}, which helps to increase support on the predictions while maintaining a high test set performance. The data-points on which we encounter high impurity values are \textbf{pruned}, i.e. the model doesn't make any prediction on those points. These data-points act as the train or test set for the next model. Finally, accuracy of all the models is only calculated for data-points on which it makes a prediction. The training procedure is described in Algorithm \ref{alg:train}:

\begin{algorithm}[!ht]
\caption{Cascading models using data pruning (training)}\label{alg:train}
\begin{algorithmic}[1]{
 \Ensure{$MaxImpurity \in (0,1)$ and $Levels\geq 0$}
 \Procedure{Cascading}{$MaxImpurity$,$Levels$, $D$}
     \State{
        $Unpruned = \phi$
     }\Comment{Contains predictions and the un-pruned data-points }
     \For{$level=1$ to $Levels$}
         \State $D' = \phi$  \Comment{Initialize an empty data-set}
         \State{model.train(D)}\Comment{Train a fresh model on the data-set}
         \For{$d \in D$}
             \State{$p = model.forward(d)$}
             \If{$GiniIndex(p) \geq MaxImpurity$}    
                \State{Append $d$ to $D'$}
            \Else\State{Append $(p,d)$ to $Unpruned$}
        \EndIf
        \EndFor
        \State{$D = D'$}
    \EndFor
    \EndProcedure
}
\end{algorithmic}
\end{algorithm}

During testing, given a sequence of models, each data-point passed through the sequence, with each model either making a prediction or passing the data-point onto the next model. The test accuracy is calculated solely on the un-pruned points. This method can be used with any classification model which can output a probability distribution over the classes. For the experiments, DDT (differentiable decision trees) and MLP were used. The inference procedure is described in Algorithm \ref{alg:infer}:

\begin{algorithm}
\caption{Inference using Cascading Models}\label{alg:infer}
\begin{algorithmic}[1]{
 \Ensure{$MaxImpurity \in (0,1)$ and $Levels\geq 0$}
 \Procedure{Predict}{$MaxImpurity$,$Models$, $D$}
     \State{
        $Unpruned = \phi$
     } \Comment{Contains predictions and the un-pruned data-points }
     \State{$n = Models.length$} \Comment{Number of models in cascading}
     \For{$d \in D$}
         \For{$i=1$ to $n$}
         \State{$model = Models[i]$}
             \State{$p = model.forward(d)$}
             \If{$GiniIndex(p) \leq MaxImpurity$}    
                \State{Append $(p,d)$ to $Unpruned$}
                \State{\textbf{break}}
        \EndIf
        \EndFor
    \EndFor
\EndProcedure
}
\end{algorithmic}
\end{algorithm}

\section{Materials: Data \& Features}
\subsection{Market Data}
We have used equity price data from the Indian equity market captured as five-minute candles in the standard open, high, low, close and volume form. Data for each ticker (stock) is normalized for each day by the close price of the first candle. Thus, each day starts with a normalised close price of one, with remaining values through the day recorded in multiples of this value. 

Further, volume data for each symbol is normalised by its historical average 5-minute volume, computed on a prior year's data, (i.e., discretisation is not based on volume values in the data itself). As a result, the time-series for each symbol-day pair are on the same scale for the purposes of training machine-learning models. 

The OHLCV values are augmented with day information, captured as the number of days since some earlier reference
date, and included as an attribute `era'.

\subsection{Synthetic Data \& Noise}
Synthetic price data is generated as sine waves, with a different frequency and amplitude for each synthetic day, or `era'.
Further, amplitude is also allowed to vary during the course of the day as a function of noise, as described later below. 

Note that each wave starts at a value of one, and a phase of zero or ninety degrees chosen randomly.
As noted earlier, base amplitude and frequency is also chosen randomly for each day. 

Next, varying levels of noise are added to each sine-wave, parameterised by two values, base-noise 
$\epsilon (t) \sim \mathcal{N}(0,\sigma)$  and peak noise computed as follows:
Each sine wave has a number of peaks, say $K$, $k = 0 \ldots K-1$. Points between each
zero-crossing are allocated to the intervening peak, defining $K$ intervals. Noise is added to each peak as peak noise 
$\epsilon_{p[k]} \sim \mathcal{N}(0,\sigma_p)$. Thus, each noisy sine wave of amplitude $A$ and $K$ peaks is characterized by two 
additional noise parameters $\sigma$ and $\sigma_p$, and computed as (for t=0 to 1, with the latter corresponding to end of day):
\[
f(t) = 1 + \epsilon (t) \pm \Sigma_{t \in I_k} A (1+\epsilon_{p[k]}) \sin (\frac{K+1}{2} \pi  t) 
\]
for $K \neq 0$. If $K=0$, a noisy ascending or descending straight line is computed as:
\[
f(t) = 1 + \epsilon (t) \pm A t
\]

Each (noisy) sine wave (i.e., for a day's worth of synthetic prices) is sampled 375 times, corresponding
to one-minute price samples. The time-series thus represents a synthetic trading day of about six and a half hours.

This series is then converted into 75 five-minute OHLC candles, and similarly normalised with the first close value as 
described earlier for market data. For synthetic data the volume is fixed at one for every candle. 

As earlier, these OHLCV values are augmented with day
information, in the `era' attribute to distinguish each random selection of frequency and base amplitude.

Note that a synthetic price dataset, as described above, is computed for a number of days (`eras') for different noise levels, and
each $(\sigma,\sigma_p)$ combination. 

\subsection{Features}
We add technical analysis features such as moving averages of price and volume, relative strength index, moving-average convergence-divergence, and Bollinger-bands to the basic OHLCV attributes. 

In addition to these standard technical indicators we add \textbf{logical} and \textbf{temporal features} as follows:
(i) Differences of selected attributes: open - close, high - low and 20 and 10 window moving averages of price. 
(ii) Slopes of close values using varying window lengths (3,5,and 10). (iii) `Change-length' for each of the
previous logical features as well as for all the base features and technical indicators: \textbf{Change-length}
is computed to be the \emph{number} of consecutive candles, i.e., current and previous five-minute
time intervals, in which a feature value is \emph{monotonically increasing or decreasing}. Change-length
is positive if the feature has been increasing and negative if it was decreasing.

Target values, i.e., values to be predicted by machine-learning, are computed as the 10-candle return, i.e.,
the difference between the normalised close value at time $t$ and time $t+50$ (since each candle represents a five minute interval).

Finally, \textit{all} the features, i.e., OHLCV values, technical indicators as well as the logical and temporal features above,
and target values are discretised into 5 bins. Thresholds for discretisation are computed using a prior year's similar data in manner so as to result in equally populated bins, i.e., a percentile-based binning. For synthetic data, an independently sampled set is used
to compute discretisation thresholds. (Thus, any potential discretisation-based information leakage is avoided during machine-learning.)

\section{Experiments \& Results}
The models, standard as well as the cascading MLPs and DDTs  are evaluated via 4 experiments for market as well as synthetic data. The experiments 
evaluate the models on their test set performance for given a combination of train and test data-set. The experiments are as follows:   
\begin{enumerate}
    \item Training on data from a set of eras and testing on data from the same set of eras.
    \item Training on data from a set of eras but testing on data from a different set of eras.
    \item Training and testing on data from a single era.
    \item Training on data from a single era but testing on data from a different era.

\end{enumerate}
In addition to the above experiments, two experiments were carried out solely on the synthetic data to examine the performance of the models when the noise levels in the train and test data-sets are different.
\begin{enumerate}
    \item Training on clean data and testing on noisy data.
    \item Training on noisy data and testing on clean data.

\end{enumerate}
K-fold cross-validation is used to determine the optimal hyper-parameters for each model. The training set used in the first experiment is used in the cross-validation for both data and the same hyper-parameters are used for all the experiments on that set of data (synthetic and market).  

\subsubsection{Cascading Models:}
Two types of models, DDT and MLP were trained using this method. DDT of depth 6 were trained for 500 epochs and MLP of hidden dimensions (128,64,32) trained for 1000 epochs were used. Additionally, for further investigation on the market data, DDTs of depth 4 trained for 1000 epochs and MLP of hidden layers (256,128,64,32) trained for 1000 epochs were also tested. For all the experiments, the maximum admissible impurity was 0.5 and the number of levels of cascading was 3. Using 3 levels, as observed in the experiments, improved the support by as much as 50\% or more compared to using a single level.
\subsection{Training on data from a set of eras and testing on data from the same set of eras}
The training data consists of the first 80\% data points from each era in the data-set. The test set consists of the latter 20\% from each era. 
These results are tabulated in Table \ref{table:se1ddt} and Table \ref{table:se1mlp}.

Unsurprisingly, performances on the test data reduce with an increase in noise. The models are highly reactive to noise addition as observed by the reduction in performance from noise level [0,0.05] to [0.01,0]. However, post this drop, even though the performance decreases, it isn't as sharp, and accuracy curves are smoother.  

The cascaded models offer tangible improvement in test set performance when compared to their single model counterparts. The Cascading MLPs provide test accuracies that are as good as, if not better, than those of Cascading DDTs albeit with a substantially lower support for both train and test sets. This result, however, is reversed for the market data, where the DDTs provide greater accuracy with lower support.

On testing the DDT of depth 4 (trained for 1000 epochs) on the market data, it was found that it gave a slightly better performance with marginally higher support. The extended MLP gives significantly better results than the one used (however the base model performance is lower), albeit with much lesser support. 

Lastly, the test-time confusion matrices revealed that the  predictions made by the cascade of models i.e., predictions on the un-pruned data, lie primarily at the extremes of the target distribution. The implications of this result are further explained in experiment 2 where a similar result was observed.

\begin{table}[!ht]
    \centering
    \begin{tabular}{|l|l|l|l|l|l|l|}
    \hline
        Noise & \multicolumn{2}{|c|}{BASE MODEL} & \multicolumn{4}{c|}{COMBINATION}  \\ \hline
       & \multicolumn{2}{|c|}{Accuracy}&\multicolumn{2}{c|}{Accuracy}&\multicolumn{2}{c|}{Support}\\ \hline
        ~ & Train & Test & Train & Test & Train  & Test \\ \hline
        [0, 0] & 0.964 & 0.964 & 0.973 & 0.972 & 0.997 & 0.993 \\ \hline
        [0, 0.05] & 0.926 & 0.866 & 0.957 & 0.897 & 0.96 & 0.954 \\ \hline
        [0.01, 0] & 0.659 & 0.604 & 0.899 & 0.808 & 0.43 & 0.42 \\ \hline
        [0.01, 0.05] & 0.71 & 0.676 & 0.858 & 0.775 & 0.677 & 0.69 \\ \hline
        [0.03, 0] & 0.554 & 0.522 & 0.83 & 0.759 & 0.317 & 0.322 \\ \hline
        [0.05, 0.05] & 0.551 & 0.485 & 0.818 & 0.726 & 0.296 & 0.288 \\ \hline
        [0.075, 0] & 0.516 & 0.47 & 0.808 & 0.706 & 0.221 & 0.224 \\ \hline
        [0.075, 0.05] & 0.546 & 0.457 & 0.81 & 0.738 & 0.237 & 0.23 \\ \hline
    \end{tabular}
      \caption{\label{table:se1ddt}Training on data from all eras and testing on data with same set of eras. (synthetic data)	 (DDT)					}
\end{table}
\begin{table}[!ht]
    \centering
    \begin{tabular}{|l|l|l|l|l|l|l|}
    \hline
        Noise & \multicolumn{2}{|c|}{BASE MODEL} & \multicolumn{4}{c|}{COMBINATION}  \\ \hline
        ~ & \multicolumn{2}{|c|}{Accuracy}&\multicolumn{2}{c|}{Accuracy}&\multicolumn{2}{c|}{Support}\\ \hline
        ~ & Train & Test & Train & Test & Train  & Test \\ \hline
       [0, 0] & 0.933 & 0.931 & 0.959 & 0.962 & 1 & 1 \\ \hline
        [0, 0.05] & 0.873 & 0.844 & 0.914 & 0.865 & 0.996 & 0.994 \\ \hline
        [0.01, 0] & 0.599 & 0.562 & 0.903 & 0.877 & 0.306 & 0.302 \\ \hline
        [0.01, 0.05] & 0.627 & 0.597 & 0.855 & 0.824 & 0.437 & 0.444 \\ \hline
        [0.03, 0] & 0.515 & 0.504 & 0.917 & 0.923 & 0.174 & 0.158 \\ \hline
        [0.05, 0.05] & 0.495 & 0.479 & 0.85 & 0.793 & 0.197 & 0.195 \\ \hline
        [0.075, 0] & 0.467 & 0.455 & 0.778 & 0.707 & 0.155 & 0.158 \\ \hline
        [0.075, 0.05] & 0.468 & 0.456 & 0.83 & 0.824 & 0.139 & 0.142 \\ \hline
    \end{tabular}
    \caption{\label{table:se1mlp}Training on data from all eras and testing on data with same set of eras. (synthetic data)	 (MLP)					}
\end{table}

\subsection{Training on data from a set of eras and testing on data from a different set of eras}
The training set consists of all the data-points from every era in a set. The test set also consists of the same, however, the eras are from a different set and there is no common era between the test and training sets. The number of eras in both sets is equal. 

\begin{figure}[!ht]
  \includegraphics[width=\linewidth]{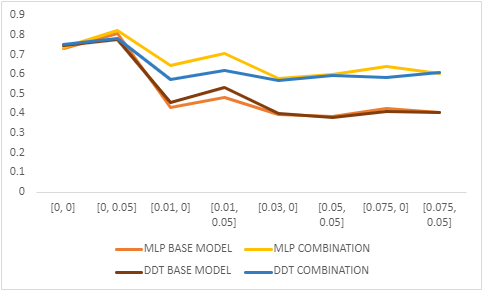}
  \caption{Test Accuracies for Base and Cascaded Models vs Noise for Experiment 2 on synthetic data}
  \label{fig:e2test}
\end{figure}

\begin{figure}[!ht]
  \includegraphics[width=\linewidth]{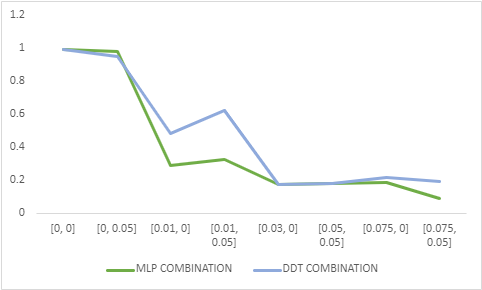}
  \caption{ Test Support vs Noise for Experiment 2 on synthetic data}
  \label{fig:e2support}
\end{figure}

The test accuracies and supports, in this case, are presented as line graphs in Figure \ref{fig:e2test} and Figure \ref{fig:e2support}.
The trend is similar to experiment 1, but the accuracy values are lower. 
Further, the accuracy level is better maintained across noise levels but leads to diminishing support at higher noise levels. On the real data (Tables \ref{table:me2ddt} and \ref{table:me2mlp}), the cascading models don't provide as much improvement in accuracy and also end up with significantly lower support. The DDT of depth 4 and the extended MLP provided improved results at the cost of lower support as shown in Tables \ref{table:me2ddtsmall} and \ref{table:me2mlpbig}. 

\begin{figure}[!ht]
  \centering
  \begin{subfigure}[b]{0.235\textwidth}
   \includegraphics[width=\textwidth]{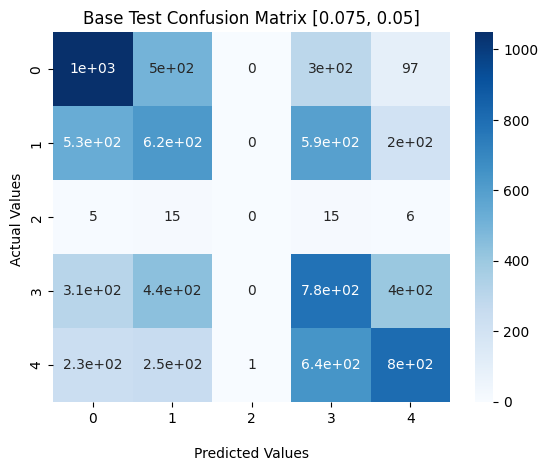}
    \caption{Base DDT}
  \end{subfigure}
  \hfill
  \begin{subfigure}[b]{0.235\textwidth}
     \includegraphics[width=\textwidth]{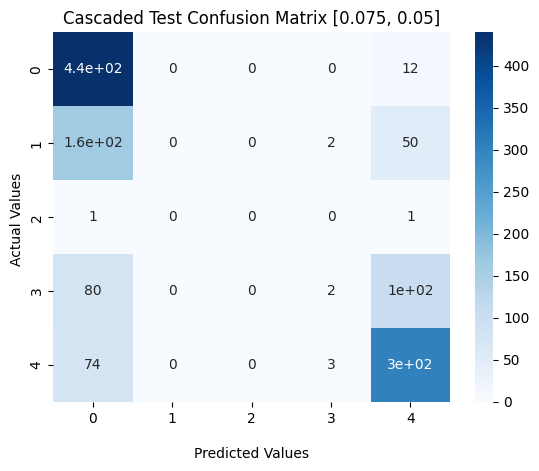}
    \caption{ Cascaded DDT}
  \end{subfigure}
  \caption{\label{figure:cmddtsyn}Confusion Matrices for DDT in Experiment 2 on synthetic data with noise level [0.075,0.05]}
\end{figure}

\begin{figure}[!ht]
  \centering
  \begin{subfigure}[b]{0.235\textwidth}
    \includegraphics[width=\textwidth]{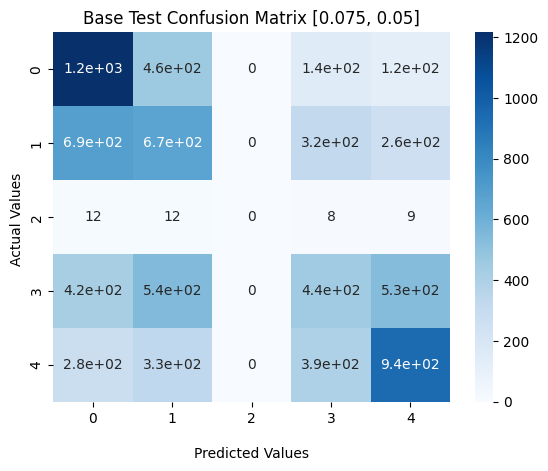}
    \caption{ Base MLP}
  \end{subfigure}
  \hfill
  \begin{subfigure}[b]{0.235\textwidth}
    \includegraphics[width=\textwidth]{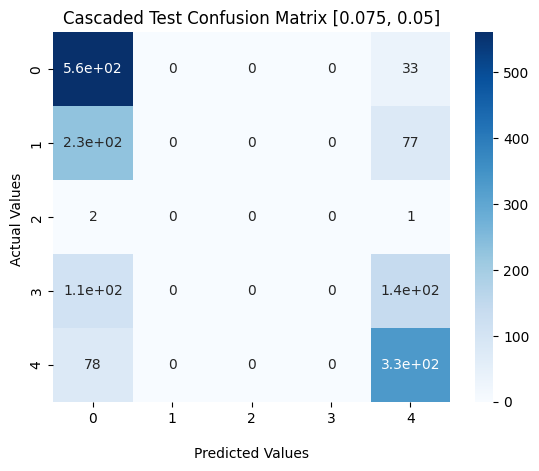}
    \caption{Cascaded MLP}
  \end{subfigure}
  \caption{\label{figure:cmmlpsyn}Confusion Matrices for MLP in Experiment 2 on synthetic data with noise level [0.075,0.05]}
\end{figure}

\begin{figure}[!ht]
  \centering
  \begin{subfigure}[b]{0.235\textwidth}
   \includegraphics[width=\textwidth]{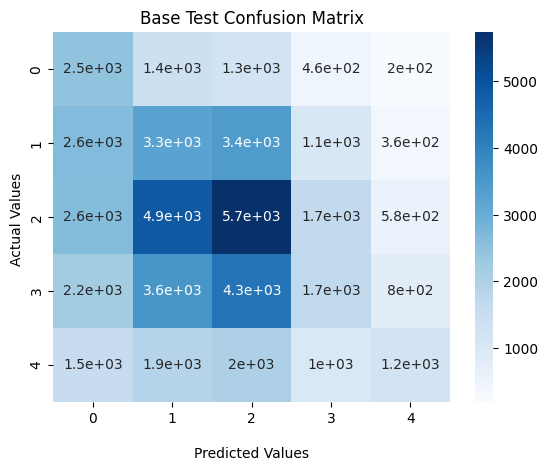}
    \caption{Base DDT}
  \end{subfigure}
  \hfill
  \begin{subfigure}[b]{0.235\textwidth}
    \includegraphics[width=\textwidth]{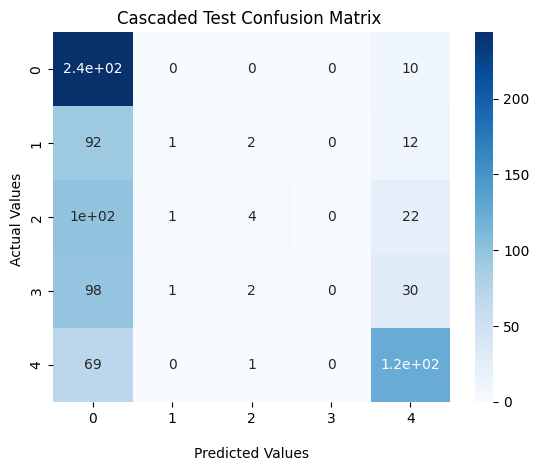}
    \caption{Cascaded DDT}
  \end{subfigure}
   \caption{\label{figure:cmddtmarket}Confusion Matrices for DDT in Experiment 2 on market data}
\end{figure}

\begin{figure}[!ht]
  \centering
  \begin{subfigure}[b]{0.23\textwidth}
   \includegraphics[width=\textwidth]{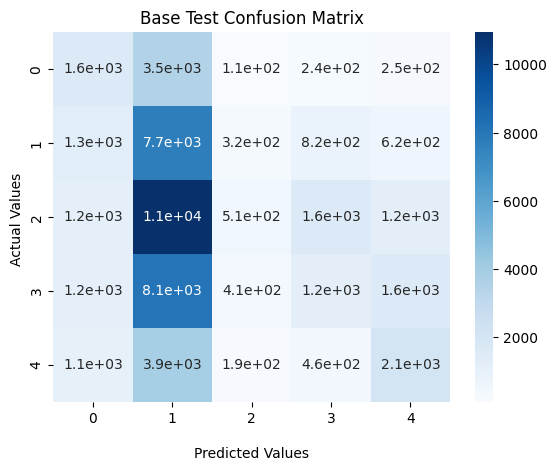}
    \caption{Base MLP}
  \end{subfigure}
  \hfill
  \begin{subfigure}[b]{0.23\textwidth}
    \includegraphics[width=\textwidth]{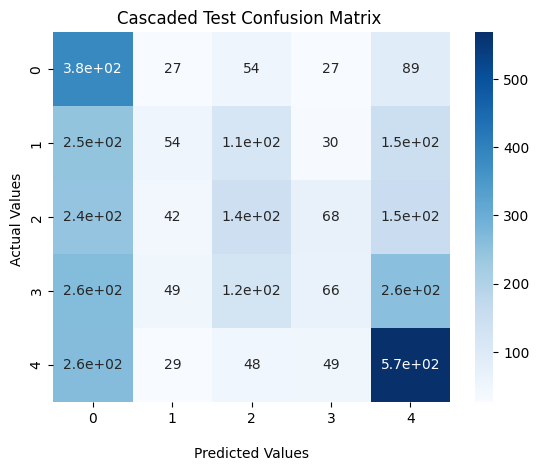}
    \caption{Cascaded MLP}
  \end{subfigure}
  \caption{\label{figure:cmmlpmarket}Confusion Matrices for MLP in Experiment 2 on market data}
\end{figure}

\textit{Most importantly}, observe Figures \ref{figure:cmddtsyn} and \ref{figure:cmmlpsyn}
that depict test-time confusion matrices for the base and cascaded models on synthetic data as well as Figures \ref{figure:cmddtmarket} and \ref{figure:cmmlpmarket} which depict the same but on market data: We observe that most of the data points on which
the cascaded models make a prediction, i.e., the un-pruned data, are at the extremes of the target distribution, i.e., predicting classes 0 or 4 as seen in experiment 1.
It is important to note that this was not inevitable. It could well have transpired that the cascaded
models only chose points in the middle of the distribution, rendering their predictions useless. As it happens, confident and accurate
predictions at the extremes are exactly what is desired for making trading decisions, i.e., buy (long) for a prediction of 4 and sell (short)
for 0. 


\begin{table}[!ht]
    \centering			
    \begin{tabular}{|l|l|l|l|l|l|}
    \hline
        \multicolumn{2}{|c|}{BASE MODEL} & \multicolumn{4}{c|}{COMBINATION} \\ \hline
         \multicolumn{2}{|c|}{Accuracy}&\multicolumn{2}{c|}{Accuracy}&\multicolumn{2}{c|}{Support}\\ \hline
        ~  Train & Test & Train & Test & Train  & Test \\ \hline
        0.392 & 0.273 & 0.798 & 0.375 & 0.053 & 0.046 \\ \hline
    \end{tabular}
    \caption{\label{table:me2ddt}Training on market data from all eras and testing on data with different set of eras. (DDT)}	
\end{table}
\begin{table}[!ht]
    \centering			
    \begin{tabular}{|l|l|l|l|l|l|}
    \hline
        \multicolumn{2}{|c|}{BASE MODEL} & \multicolumn{4}{c|}{COMBINATION} \\ \hline
        \multicolumn{2}{|c|}{Accuracy}&\multicolumn{2}{c|}{Accuracy}&\multicolumn{2}{c|}{Support}\\ \hline
        ~  Train & Test & Train & Test & Train  & Test \\ \hline
       0.333 & 0.265 & 0.802 & 0.475 & 0.026 & 0.022 \\ \hline
    \end{tabular}
     \caption{\label{table:me2ddtsmall}Training on market data from all eras and testing on data with different set of eras. (DDT with depth 4 and trained for 1000 epochs)}	
\end{table}
\begin{table}[!ht]
    \centering		
    \begin{tabular}{|l|l|l|l|l|l|}
    \hline
        \multicolumn{2}{|c|}{BASE MODEL} & \multicolumn{4}{c|}{COMBINATION} \\ \hline
        \multicolumn{2}{|c|}{Accuracy}&\multicolumn{2}{c|}{Accuracy}&\multicolumn{2}{c|}{Support}\\ \hline
        ~  Train & Test & Train & Test & Train  & Test \\ \hline
      0.314 & 0.25 & 0.847 & 0.335 & 0.067 & 0.055 \\ \hline
    \end{tabular}
     \caption{\label{table:me2mlp}Training on market data from all eras and testing on data with different set of eras. (MLP)}		
\end{table}
\begin{table}[!ht]
    \centering			
    \begin{tabular}{|l|l|l|l|l|l|}
    \hline
        \multicolumn{2}{|c|}{BASE MODEL} & \multicolumn{4}{c|}{COMBINATION} \\ \hline
        \multicolumn{2}{|c|}{Accuracy}&\multicolumn{2}{c|}{Accuracy}&\multicolumn{2}{c|}{Support}\\ \hline
        ~  Train & Test & Train & Test & Train  & Test \\ \hline
     0.314 & 0.243 & 0.783 & 0.461 & 0.062 & 0.046 \\ \hline
    \end{tabular}
     \caption{\label{table:me2mlpbig}Training on market data from all eras and testing on data with different set of eras. (MLP with an extra layer)}	
\end{table}
\subsection{Training on a single era and testing in the same era}
\label{sec:single_era_same}
The training set in this experiment is the first 80\% of the data from a single era and the test set is the remainder of the data. Performances on eras are averaged over the results on a noise level. 

Due to lower data-set sizes, the models tend to over-fit the data, with test accuracy on the market data coming to be about 30\%, which is much lower than the train accuracies. 

The Cascading models also don't provide much improvement, as the test accuracies on both, synthetic and market data, are almost the same as that of the base model. Due to over-fitting, the model gives low impurities for most of the points as observed by the supports, while pruning, it does not extract the data-points which is more helpful for increasing test accuracy. 

\subsection{Training on a single era and testing on a different era}
\label{sec:single_era_diff}
In this experiment, O(N) combinations of train and test era are used, where N is the number of eras in each of the two sets in experiment 2. The entirety of the data from the eras is used in the train and test sets. While training performances are usually good, the test accuracy usually depends on the similarity in distribution between the data in the two eras.
\subsection{Performance in cases where train and test set have different noise levels}
\label{sec:diff_noise}
The following experiments were conducted only on the synthetic data to assess the ability of models to extract patterns from the training data and effectively apply it to data at a different noise level. 
\subsubsection{Training on clean data and testing on noisy data}
In this case, the models give low accuracies once noise is added and remain approximately the same for all noise levels. The cascading models don't improve upon this either. For the MLP, the mean test accuracy is 32.9\% and lies in the range (30.8,34.3). For DDTs, the mean is slightly higher at 37.5\% and lies in a longer range of (35.1,44). This is due to the model over-fitting on the clean data, which forces it to give predictions with low impurity even with very noisy data which hampers the working of the cascading model.
\subsubsection{Training on noisy data and testing on clean data}
For the standard ensemble models and the base models, the training accuracy decreases with increasing noise, and the test accuracy remains significantly lower than the train. However, the cascading models offer significant improvement in this case.  Here,  the Cascading MLPs give significantly better results than the base models, and better results than the Cascading DDTs, which give a lower accuracy but with higher support.

We omit the detailed results for the above cases in Sections \ref{sec:single_era_same}-\ref{sec:diff_noise} due to lack of space.

\subsection{Utility and Risk-adjusted Return}
The confusion matrices formed by the cascading models in experiments 1 and 2 reveal that a major portion lies at the extremes of the target distribution, 0 and 4. These 2 classes generally determine the trading decision to be made at a certain point in time and are highly actionable, making these predictions very useful. 

We use a metric to compare the average gain per prediction (utility) of the models before and after cascading. This value is calculated using the points where the predictions are either 0 or 4 (extreme ends of the classes). Since the `target' refers to stock-specific returns, decisions to trade
would mainly take place when a target of 0 or 4 is predicted, so we focus only on these two columns. Each pair of predictions and ground truth has a `utility' attached to it. If the ground truth is 2, the utility is 0 regardless of the prediction as there would be no profit/loss in any case. For predictions that are 0, if the ground truth is 0, a utility of +2 is awarded, indicating a successful short sale. If the ground truth is 1, an utility of +1 is awarded, since a short sale would still have been profitable, albeit less so. For ground truths of 3 and 4, -1 and -2 are awarded to the model as the prediction would result in a loss. Predictions on class 4 award reverse. Then:
\[Average Utility = \frac{(Gain-Loss)}{\#~of~predictions~of~class~0~or~4}\]
In addition to this, the downside-risk-adjusted-return is also calculated for each model, to measure the gain vs
risk to capital: 
\[DRAR = \frac{(Gain-Loss)}{Loss}\] 
as well as a `Traded Sharpe ratio', as the ratio of utility to the standard deviation of returns \textit{measured
across the trades actually recommended, i.e., when class 0 or 4 is predicted}. We motivate utility, DRAR, and Traded Sharpe metrics further in the next Section below.

Table \ref{table:utils} reports utility, DRAR, and Traded Sharpe for experiments 1 and 2, and for synthetic and market data.
\textbf{Cascaded DDT performs better than the base model in all cases, and also better than the cascaded
MLP}, except for the case of market data in experiment 1. (This may be because the higher capacity MLP captures more era-specific patterns).

\begin{table*}[!ht]
    \centering
    \begin{tabular}{|l|l|l|l|l|l|}
    \hline
        \multicolumn{6}{|c|}{Experiment 1} \\ \hline
        Data & Metric& \multicolumn{2}{c|}{DDT} & \multicolumn{2}{c|}{MLP} \\ \hline
        Dataset & ~ & Base & Cascaded & Base & Cascaded \\ \hline
        \multirow{4}{*}{Synthetic ([0.075,0.05]) - 1571 points} & Average Utility & 1.11 & \textbf{1.69} & 1.06 & 1.50 \\  \cline{2-6}
        ~ & Gain-Loss & 1073-202 & 303-4 & 1133-229 & 488-42 \\ \cline{2-6}
        & Downside-risk adjusted return & 4.31 & \textbf{74.7}5 & 3.95 & 10.62 \\ \cline{2-6}
        &Traded Sharpe ratio  &0.92 & \textbf{2.93} & 0.88 & 1.53 \\ \hline
        \multirow{4}{*}{Market-4522 points} & Average Utility &0.50 & 1.29 & 0.79 & \textbf{1.46} \\  \cline{2-6}
        ~ & Gain-Loss & 1590-683 & 175-25 & 1227-327 & 107-9 \\ \cline{2-6}
        & Downside-risk adjusted return &  1.33 & 6 & 2.75 & \textbf{10.89} \\ \cline{2-6}
        &Traded Sharpe ratio &0.38 & 1.07 & 0.63 & \textbf{1.46} \\ \hline
        \multicolumn{6}{|c|}{Experiment 2} \\ \hline
        Data & Metric& \multicolumn{2}{c|}{DDT} & \multicolumn{2}{c|}{MLP} \\ \hline
        Dataset & ~ & Base & Cascaded & Base & Cascaded \\ \hline
        \multirow{4}{*}{Synthetic ([0.075,0.05]) - 7729 points} & Average Utility & 0.94 & \textbf{1.18} & 0.90 & 1.12 \\  \cline{2-6}
        ~ & Gain-Loss & 4530-1164 & 1740-302 & 5500-1480 & 2150-409 \\ \cline{2-6}
        & Downside-risk adjusted return &2.89 & \textbf{4.76} & 2.72 & 4.26 \\ \cline{2-6}
        &Traded Sharpe ratio & 0.73 & \textbf{0.98} & 0.70 & 0.92\\ \hline
        \multirow{3}{*}{Market-55450 points} & Average Utility & 0.33 & \textbf{0.72} & 0.47 & 0.50 \\  \cline{2-6}
        ~ & Gain-Loss & 10800-5960 & 842-268 & 10300-4520 & 2410-1108 \\ \cline{2-6}
        & Downside-risk adjusted return & 0.81 & \textbf{2.14} & 1.28 & 1.18 \\  \cline{2-6}
        &Traded Sharpe ratio& 0.26 & \textbf{0.54} & 0.37 & 0.36\\ \hline
    \end{tabular}
    \caption{Average Utilities, Gain-Loss, Downside-risk-adjusted-returns and Traded Sharpe ratios for Experiment 1 and 2}\label{table:utils}
\end{table*}

\section{Discussion}
Consider an algorithmic trading scenario where a machine-learning model is
being used to recommend long/short trades based on its prediction of where the
market will be after a fixed number of time-steps. 
Consider two strategies: (a) using a model that recommends only a small number of trades each with high utility,
i.e., expected gain, and (b) using a model that recommends a large number of trades but each having lower utility.

While it is conceivable that the expected total return is higher for strategy (b), as can also be seen from
the figures in Table \ref{table:utils}, in the world of financial trading minimising risk is as important as maximizing
return. The standard mechanism for measuring risk-adjusted return is the Sharpe ratio, which measures the overall
expected gain per time-step adjusted by the standard deviation of returns. However, since our models predict far
less often than the base models, we instead measure downside-risk adjusted return and Traded Sharpe ratio as defined earlier.
As we have already seen, the cascaded models, especially the DDT-based ones, are far superior
in terms of these metrics as reported in Table \ref{table:utils}.

To further motivate these metrics, consider that short-term trading often takes place in a leveraged manner, i.e.,
using funds borrowed from a broker and only minimal capital of the trader's own; in practice this ratio is at least
5X and can be as high as 10X.

A strategy that incurs large draw-downs (i.e., intermediate losses)
risks margin calls in the process, requiring the traders to put up extra capital that could
even wipe them out if they are operating with high leverage.

On the other hand, strategies that avoid large losses at all costs, including lower overall returns, allow
traders to put up more base capital and/or take higher leverage and thus increase their total return, 
with the risks of serious margin calls that could wipe them out being far reduced. 

\section{Related Work}
\label{related_work}
Applying machine learning to financial markets has been and is probably continuing to be
widely attempted in practice as well as in research. Apart from application of the standard
data-science pipeline using off-the-shelf models, which we do not recount, some novel approaches
deserve mention. 

A number of prior works such as \cite{dhar2019transforming} and \cite{cohen2020trading} have converted financial data into image data and thereafter applied CNN-based deep-learning models for predicting returns. While the latter directly use images
of price-series, the former converts level-2 data, i.e., the limit-order book over a window into an image.

Recent works applying machine learning to trading based on price signals alone have also used reinforcement learning:
`Deep Momentum Networks' \cite{lim2019enhancing} as well as `Momentum Transformer'
\cite{wood2021trading} formulate the trading task as one of suggesting the position to take, e.g., 1 for a long
(i.e., buy) position, -1 for a short (i.e., sell) position, and 0 for no position. Exiting a buy/sell position 
takes place when a 0 action follows the previous 1 actions, etc. Neural networks are trained to directly
optimize volatility-adjusted expected returns, adjusted for transaction costs, over a trading period. 
The former paper uses MLPs and LSTMs, while the latter uses transformers.
Both works are essentially applying vanilla REINFORCE to the MDP formulation of the trading problem.
Deep Reinforcement Learning in Trading \cite{zhang2020deep} uses the same formulation as the above two works,
but applies more refined reinforcement learning techniques, e.g., policy-gradient, actor-critic, and deep-Q-learning algorithms.
Most recently, meta-reinforcement learning via the RL$^2$ algorithm \cite{duan2016rl} was used along with logical features (including temporal ones) learned automatically via ILP \cite{harini2023neuro}.

The case for using logical rule-based models for financial data as an antidote to noise was made in \cite{dhar2011prediction}.
More generally, the original CN2 algorithm \cite{clark1989cn2} was developed explicitly to deal with noisy data;
RIPPER \cite{asadi2016ripmc} is a modern version of the same.
More modern approaches to combat noise, specifically label noise (i.e., features are assumed relatively noise-free 
in comparison) are represented by \cite{patrini2017making} and \cite{qin2019making}. These 
use variations on novel loss functions or replace
targets of suspected noisy samples with model predictions during training (`boot-strapping'). 

Dealing with noise is related to works on calibration, i.e., ensuring that predicted probabilities are close to
observed frequencies on train and test datasets. One such recent approach also uses data pruning during training
(as opposed to post-training, as in our proposed approach),
to improve calibration error as well as improve fraction of high-confidence predictions \cite{patra2023calibrating}.

\section{Conclusions and Future Work}
We have introduced cascading models as an approach to deal with noisy data and make
predictions only on the fraction of data where the model cascade is confident. We
have presented results using cascaded differentiable decision trees as well as MLPs
on synthetic data with varying levels of noise as well as real market data. 

We observe that using cascaded models results in more accurate
predictions and degrade more gracefully with noise at the expense of support.
Further, the predictions that \textit{are} made
have high utility towards making trading decisions, as well as result in better risk-adjusted
returns as per the approximate metric used.
We also find that the cascaded differentiable decision trees perform better than 
cascaded MLPs, especially in the utility of their predictions and in terms of risk-adjusted returns
and traded Sharpe ratio.

Performances of all the models evaluated degrade when tested on distributions different from those they are trained on (represented by
different eras in our case). Future work towards addressing this could be to employ meta-learning
\cite{finn2017model} or continual learning \cite{finn2019online} techniques, combining these with the idea of
using cascaded models via data-pruning as we have proposed. Additionally, cascaded models may use train-time pruning approaches such as in \cite{patra2023calibrating}.

%
\bibliographystyle{ACM-Reference-Format}
\bibliography{sample-base}


\end{document}